\documentclass{article}
\usepackage{spconf,amsmath,graphicx}
\usepackage{hyperref}
\usepackage{algorithm}
\usepackage{algorithmic}

\makeatletter
\def\@name{ \emph{T. Syeda-Mahmood, H. Ahmad, N. Ansari, Y. Gur,  S. Kashyap, A. Karargyris,} \\ 
\emph{M. Moradi, A. Pillai, K. Sheshadri , W. Wang, K. C. L. Wong,  J. Wu} \\
\emph{} 
}
\makeatother

\title{Building a Benchmark Dataset and Classifiers for Sentence-Level Findings in AP Chest X-rays
\thanks{This paper was accepted by IEEE ISBI 2019. \copyright 2019 IEEE. Personal use of this material is permitted. Permission from IEEE must be obtained for all other uses, in any current or future media, including reprinting/republishing this material for advertising or promotional purposes, creating new collective works, for resale or redistribution to servers or lists, or reuse of any copyrighted component of this work in other works.}
}
\address{IBM Almaden Research Center\\
contact:stf@us.ibm.com}
\begin{document}
%\ninept
%
\maketitle
\begin{abstract}
Chest X-rays are the most common diagnostic exams in emergency rooms and hospitals.
There has been a surge of work on automatic interpretation of chest X-rays using deep learning approaches after the availability of large open source chest X-ray dataset from NIH. However, the labels are not sufficiently rich and descriptive for training classification tools.  Further, it does not adequately address the findings seen in Chest X-rays taken in anterior-posterior (AP) view which also depict the placement of devices such as central vascular lines and tubes.  In this paper, we present a new chest X-ray benchmark database of 73 rich sentence-level descriptors of findings seen in AP chest X-rays. We describe our method of obtaining these findings through a semi-automated ground truth generation process from crowdsourcing of clinician annotations. We also present results of building classifiers for these findings that show that such higher granularity labels can also be learned through the framework of deep learning classifiers. \end{abstract}
%based on pure deep learning and hybrid deep learning formulations where the feature representation learning of deep learning networks is combined with the feature selection and tuning of ensemble classifiers.
%As the multi-class classification within deep learning networks often implicitly assumes mutually exclusive labels,
\vspace{-0.1in}
\section{Introduction}
\label{sec:intro}
\begin{figure}
	\centering
	\includegraphics[width=8cm]{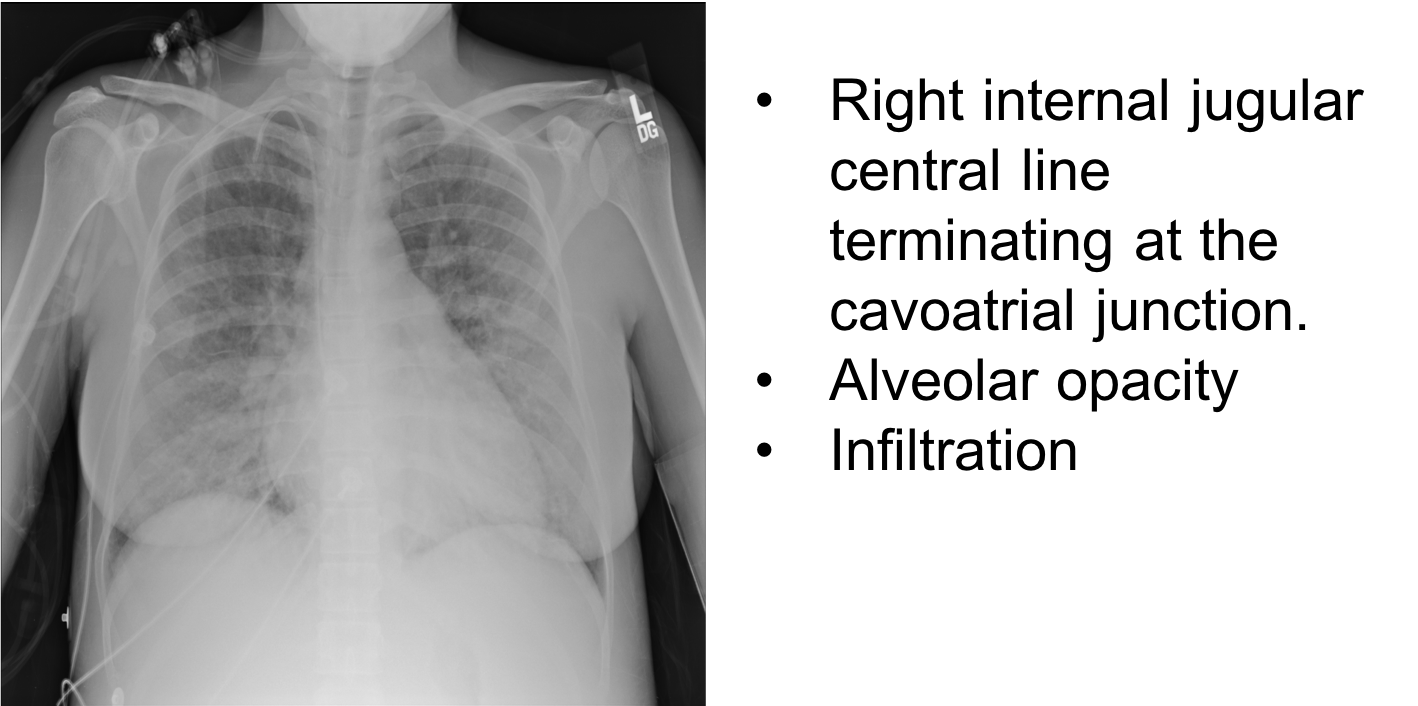}
	\caption{Illustration of spatial overlap between labels relating to lines and tubes and anatomical findings.}
	\label{spatialoverlap}
\end{figure}

Chest X-rays are the most common imaging exams being conducted in emergency rooms. Recently, a number of researchers have begun automated interpretation of chest X-rays, focusing on posterior-anterior (PA) views and limited number of labels of high granularity such as opacity or consolidation.\cite{rajpurkar2017chexnet,wang2017chestx,laserson2018textray}. If machines are to assist radiologists through automated interpretation, it is important to expand the number of findings as well as refine them to incorporate location, laterality, character and other information so that automated report generation may one day become possible.  Further, the viewpoints should be expanded to cover anterior-posterior (AP) views as well which are generally taken to aid in the diagnosis of acute and chronic conditions in intensive care units in hospitals. Although the AP view is lower in quality to PA view, this is often the only mode in which sick patients can be imaged for problems in lungs, bony thoracic cavity, mediastinum, and great vessels. The resulting images often depict multiple types of findings such as anatomical findings, technical assessment problems, and tubes/lines placement issues as shown in Figure~\ref{spatialoverlap}.

Currently no labeled datasets that cover the list of possible findings seen in AP chest X-rays.
The large open source chest X-ray dataset provided from NIH\cite{wang2017chestx} covers only discrete anatomical findings. Hence existing approaches have either ignored the viewpoint during training \cite{rajpurkar2017chexnet,wang2017chestx} or focused on PA views only\cite{laserson2018textray}.  %Finally, there is large imbalance in the number of training images that depict specific combinations of findings making it difficult to build robust classifiers.

 %placement of the tubes/lines at different locations where the labels refer to the device present but as different assessing the positional differences in the two semantically similar labels of "left picc (peripherally inserted catheter) lines" but at different anatomical positions as indicated in their labels.
\begin{figure}
	\centering
	\includegraphics[width=8cm]{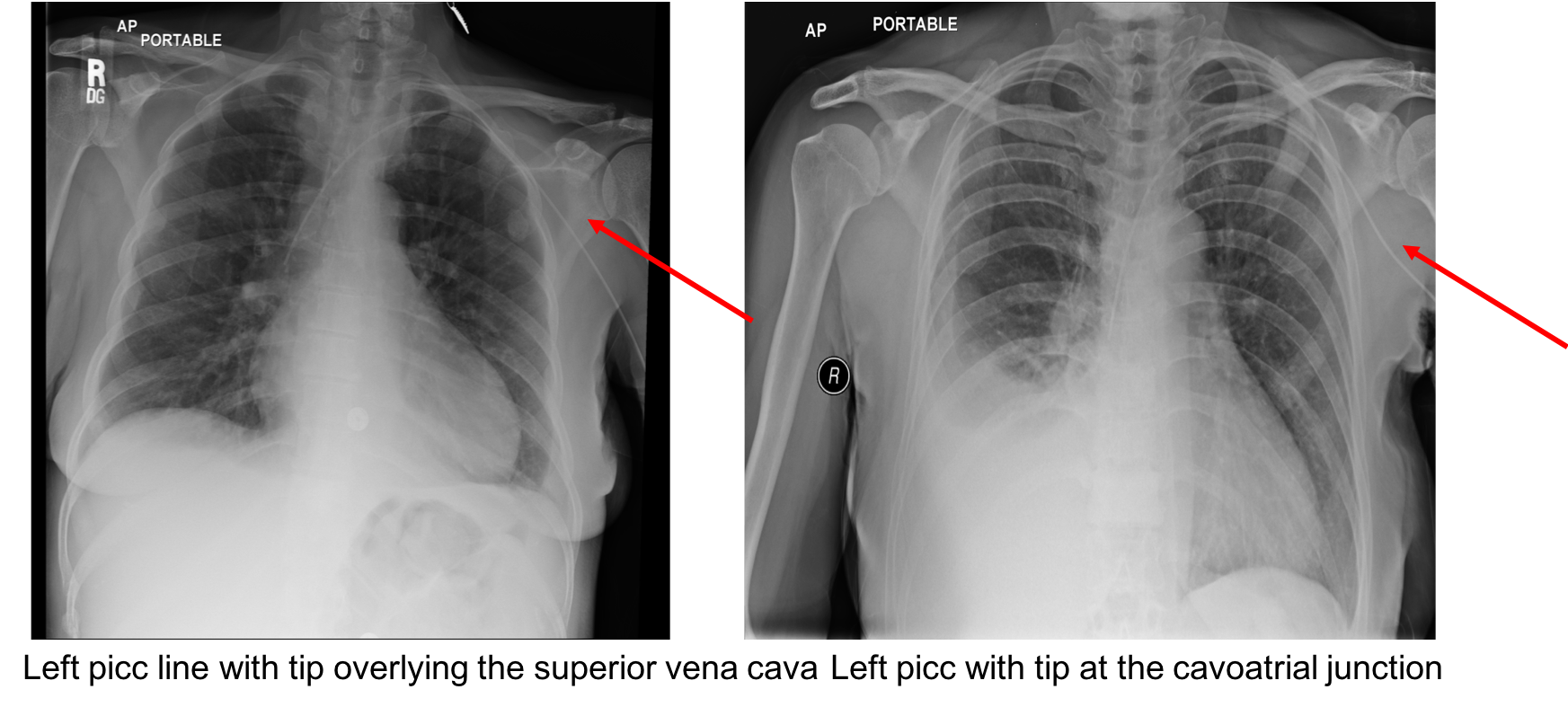}
	\caption{Illustration of semantic overlap in labels both referring to different positioning of left picc lines.}
	\label{semanticoverlap}
\end{figure}

Chest X-rays also depict spatially and semantically overlapping findings for which simple labels describing only the core finding are not sufficient to build robust classifiers. Figure~\ref{spatialoverlap} illustrates the difficulty of detecting spatially overlapping findings in AP chest X-rays. Here, the findings of "infiltration", "alveolar opacity", and positioning of "right internal jugular line at the cavoatrial junction" are all within the same spatial vicinity.  When feature regions of two different class labels spatially overlap, the classifiers often select weights that bias towards the label with larger training data. Figure~\ref{semanticoverlap} shows the specificity of labels problem with AP chest X-rays where semantic overlap between findings need to be carefully distinguished by describing not only the core findings but also the placement. Here the classification method needs to be fine grained to recognize the difference in the similarity of the labels referring to the same device but placed at different locations in the body. This requires higher precision in deriving the labels themselves.

In the work on chest X-rays so far, not enough attention has been paid to the choice of labels for the classifiers. Although about half of the NIH dataset consists of AP chest X-rays (44,812 images), the labels currently provided only cover anatomical findings without elaborating on modifiers such as laterality, location or severity which affect visual appearance and hence the classification accuracies.  Fresh annotation efforts have also begun but focused on a single finding such as the pneumonia dataset recently provided by RSNA-Kaggle challenge\cite{rsnakaggle}. Since all AP chest X-ray findings are also documented in radiology reports, automatic interpretation algorithm must support AP chest X-ray imaging. Hence there is a need to derive higher granularity labels for AP Chest X-ray imaging.
%to develop classifiers covering these images and spur the next generation work in automated chest X-ray interpretation.

 %However, this dataset refers to discrete anatomical findings only without elaborating on modifiers such as laterality, location or severity which affect visual appearance.Further, it does not adequately address the findings seen in Chest X-rays taken in anterior-posterior (AP) view.
% Several methods have attempted automated recognition of findings using multi-class classifiers built with deep learning layers for feature extraction and fully connected layers for multi-class classification on this dataset \cite{rajpurkar2017chexnet,wang2017chestx}.
 \begin{figure}
	\centering
	\includegraphics[width=8cm]{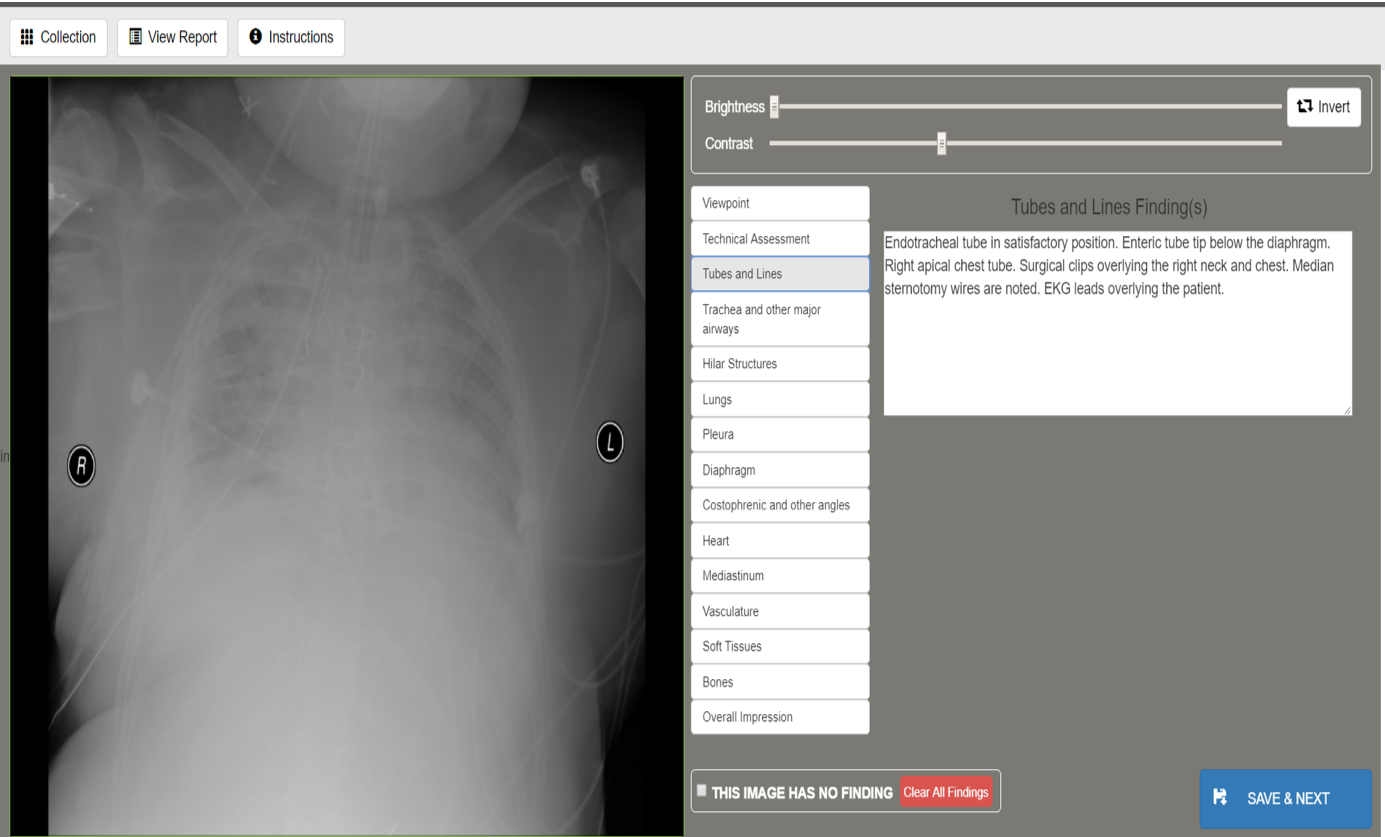}
\caption{Illustration of structured template used to collect radiology report to derive the labels. }
	\label{template}
\end{figure}

In this paper we present a benchmark dataset of AP chest X-rays originally derived from NIH dataset but relabeled for higher granularity findings in AP chest X-rays. The labels are derived from a semi-automatic curation process that involves crowd-sourcing of clinician read reports through ease-of-use user interfaces followed by automated text clustering and semantic grouping, and final clinician verification. To obtain sufficient granularity of description, sentence level labels are retained. Such labels also facilitate the production of automated reports as they can be directly used to form the report.  We also address the problem of building classifiers for such higher granularity findings by exploring two architectures, one based on conventional deep learning and another hybrid deep learning formulation to exploit the greater feature selection and explain capability of ensemble classifiers. These results show that such higher granularity findings in AP chest X-rays can also be learned by the state-of-the-art classifiers.

%We then explore the recognition of findings using a  new deep ensemble network (DBEN) that combines the advantages of feature representation learning of deep learning networks with the feature selection and tuning shown by ensemble classifiers to yield improvement in the classification of overlapping findings in AP chest X-rays.
%In this paper, we present a new hybrid deep learning framework called the Dense Binary Ensemble (DBE) which combines the rich feature extraction capabilities of deep learning networks with the one-vs-all classification capabilities of binary ensemble classifiers such as random forests.
\begin{figure}
	\centering
	\includegraphics[width=8cm]{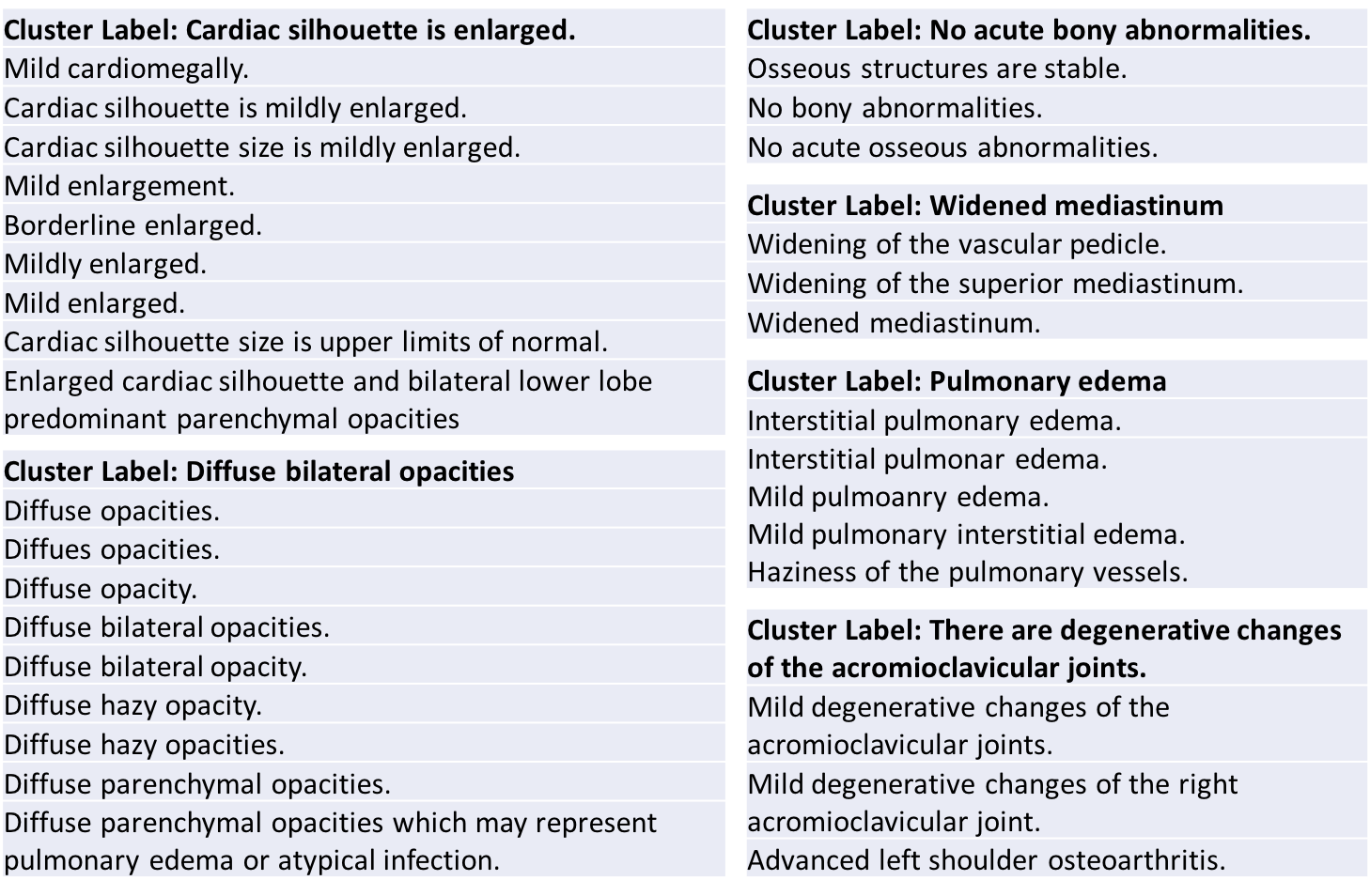}
\caption{Illustration of results of semantic clustering of report sentences. }
	\label{clusters}
\end{figure}
\begin{figure}
	\centering
	\includegraphics[width=8.5cm]{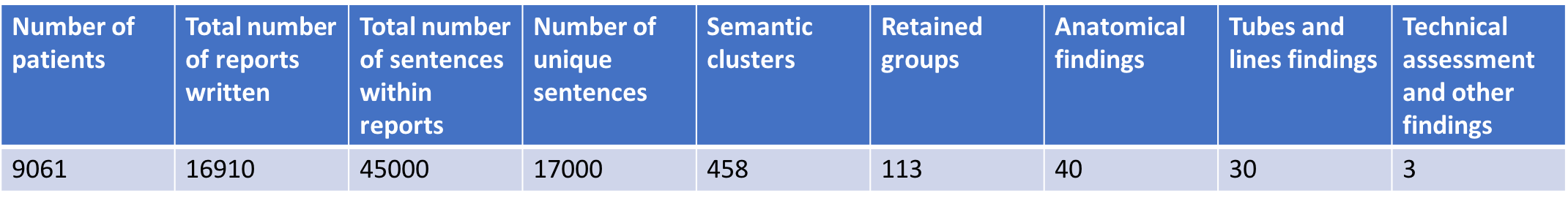}
\caption{Illustration of label data reduction through successive processing. }
	\label{tablefindings}
\end{figure}
\vspace{-0.2in}
\section{Label generation for AP Chest X-rays}
 The NIH dataset consists of 112,121, images  with 44,812 images in AP view from 9061 patients.  To generate the labeled dataset for AP chest X-rays, we sampled the dataset so that at least one AP chest X-ray image was selected from all patients to obtain a total of 16910 unique images for re-annotation. To rapidly annotate the large dataset and list of possible findings in AP chest X-rays,  we developed a web-based crowd-sourcing annotation system and recruited over 42 radiologists around the country to perform the annotation. In collecting the annotation, we simulated a radiology read setup in a hospital by providing a templated report form to the radiologists shown in Figure~\ref{template}. As can be seen by the template, it allows radiologists to describe all major structures seen in chest X-rays as well as any device artifacts including lines and tubes. Further, technical assessment was also captured in addition to structured labeling of viewpoints. The free text form within these templated sections allowed radiologists freedom to describe or dictate the findings relatively freely to rapidly complete a report (20 images/per hour was the observed speed) without requiring the selection of discrete labels.
 \vspace{-0.1in}
\subsection{Sentence clustering for label generation}
The resulting reports generated nearly 45000 sentences of which 17000 unique sentences were discovered after normalization by removing the stop words, small typos, case differences, etc. while still maintaining the order of the words. Clustering was then attempted within sentences coming from the same report template section. The distance metric chosen for clustering measured the extent of overlap of words between two sentences with and without keeping the order of the words. The pairwise similarity between two sentences $S=<s_{1}s_{2}...s_{K}>$ of K words, and $T=<t_{1}t_{2}...t_{N}>$  of N words was defined as:
\begin{equation}
d(S,T)=max\{d_{unordered}(S,T),d_{ordered}(S,T)\}
\end{equation}
\noindent where $d_{unordered}(S,T)$ is given by
\begin{equation}
d_{unordered}(S,T)=\frac{2*|S\cap T|}{|S\cup  T|}
\end{equation}
\noindent where $|S\cap T|$ is the number of words common between $S$ and $T$ and $|S\cup T|$ is the total length of the two strings in words.

\noindent The ordered score  $d_{ordered}(S,T)$ is the ordered similarity computed by a string matching algorithm called the longest common subfix (LCF) algorithm \cite{syeda2015learning} given by
$d_{ordered}(S,T)=<p_{1}p_{2}...p_{L}>$ , where L is the largest subset of words from S that found a partial match in T and $p_{i}$  is a partial match of a word $s_{i}\in S$  to a word in T. A word $s_{i}$ in S is said to partially match a word $t_{j}$ in T if it shares a maximum length common prefix  $p_{i}$ such that  $\frac{|p_{i}|}{\mbox{max}\{|s_{i}|,|t_{j}|\}}>\tau$.  If we make the threshold $\tau=1.0$ , this reduces to the case of finding exact matches to words of S. Note that this formulation is different from the conventional longest common subsequence (LCS) string matching as there is an emphasis on character grouping into words and the use of word prefixes to relate words in the English language. This algorithm uses dynamic programming alignment at the words level using word prefixes and allows for gaps and insertions while preserving the word order.  The algorithm also uses other enhancements for negation pattern finding, and abbreviation expansions as described in  \cite{syeda2015learning}.

The ordered score  $d_{ordered}$  can be computed using dynamic programming alignment algorithm by keeping an array $C[i,j]$ to calculate the score of matching a fragment of $S$ up to the ith word and fragment of $T$ up to the jth word. We then update the dynamic programming matrix according to the algorithm shown in Algorithm $1$. Here $p_{max}(i,j)$ is the longest prefix of the strings $(s_{i},t_{j})$   and  $\delta$ is a mismatch penalty, which controls the separation between matched words and prevents words that are too far apart in a sentence from being associated with the target sentence.  Using this algorithm, $S$ is said to match sentence $T$ if $\frac{|LCF(S,T)|}{|S|}\ge\Gamma$ for some threshold $\Gamma$. The choice of $\delta$ and $\Gamma$ affect the closeness of the match and were chosen to meet specified criteria for precision and recall based on an ROC curve analysis on labeled collection.

Figure~\ref{clusters} shows the results of applying the similarity score $d(S,T)$ on a variety of sentences found in the generated reports. It can be seen that the algorithm spots sentences with similar meanings without a deep understanding of their linguistic origins. The algorithm uses other enhancements for handling negations and abbreviation expansions which are skipped here for brevity.
\vspace{-0.1in}
\begin{algorithm}\label{alg1}
\small \caption{Longest Common Subfix Algorithm} \textbf{LCF(S,T):}
\begin{algorithmic}[1]
\STATE \textbf{Input/Output} Input: two strings (S,T). Output: an alignment score.
\STATE \textbf{Initialize} $c[i,0]=0,c[0,j]=0,0\leq i\leq K, 0\leq j\leq N$.
\STATE \textbf{Iterate} for $(1\leq i \leq K)$\\
                \hspace{0.4in} for $(1\leq j\leq N)$\\
\hspace{0.6in}$\rho_{ij}=\frac{|p_{max}(i,j)|}{\mbox{max}\{|s_{i}|,|t_{j}|\}}$\\
\hspace{0.6in} If $C[i-1,j-1]+\rho_{ij}>C[i-1,j]$ and \\
\hspace{0.8in} $C[i-1,j-1]+\rho_{ij}>C[i,j-1]$\\
\hspace{1in} $C[i,j]=C[i-1,j-1]+\rho_{ij}$\\
\hspace{0.6in} else\\
\hspace{0.8in}  if $(C[i-1,j]+\rho_{ij}>C[i,j-1]$\\
\hspace{1in} $C[i,j]=C[i-1,j]-\delta$\\
\hspace{0.8in} else\\
\hspace{1in} $C[i,j]=C[i,j-1]-\delta$
\end{algorithmic}
\end{algorithm}

To perform clustering, all unique sentences belonging to a section heading across reports are collected and lexicographically ordered. Starting from the first sentence, each successive sentence is added to the cluster if its LCF distance is within a threshold with respect to all previous members. The first sentence that violates this constraint becomes the start of a new cluster. This method of grouping brings out the lexical similarity in the sentences as shown in Figure~\ref{clusters}. Here a representative sentence from that group is used to denote the cluster. Using this process,  the total sentences to examine reduced from 40,000 to about 458 cluster representatives as shown in Figure~\ref{tablefindings} based on the number of clusters produced (also 458).  The semantic merging of these labels is then done manually by radiologists on this reduced dataset to further group the labels into 113 semantic groups. In doing the grouping, the radiologists kept the distinction of location, laterality and severity as those cause changes in visual appearance. By retaining all those clusters with more than 50 images per cluster, we retain 73 labels as important labels for AP chest X-rays. Looking at the distribution of labels in Figure~\ref{tablefindings} and Table~\ref{results}, we can see that there are labels related to tubes and lines, not previously known to researchers working with the NIH dataset.  %In this paper, we focus on the recognition of multiple overlapping findings in AP chest X-rays. Overlapping findings have been addressed so far using the familiar multi-class framework in deep learning networks using softmax or other output label generation strategies. When the finding labels are not mutually exclusive, there is no particular advantage to using a multi-class over a one-vs-all framework, which have been attempted before in traditional machine learning classifiers\cite{svm,rf}. However, attempting one-vs-all will be prohibitive in the case of deep learning networks when the number of classes grow. In this paper, we present a new deep learning framework called the Dense Binary Ensemble (DBE) which combines the rich feature extraction layers of Densenets with the one-vs-All classification capabilities of binary ensemble classifiers. Specifically, we propose to replace the last fully-convolutional multi-class classification layer in a fully-trained DenseNet with a N-binary ensemble classifiers each trained per overlapping class using the features derived from Densenet feature extraction layer. The new network is shown to outperform multi-class DenseNet in terms of average AUC performance across the classes trained.

%To form the labels we first performed pairwise adopted a clustering algorithm that alphabetically sorted the sentence using lexicographic ordering. The highest ranking pairs were then used in a clustering algorithm that ensured that every member of the cluster was within a distance lower than a threshold with respect to the pairwise similarity score. The resulting clusters were then examined for semantic accuracy by the radiologists and a representative sentence from the cluster was picked by the radiologists as a label for the cluster. Figure~\ref{clusters} shows clusters of sentences that were all grouped under a common semantic label. clusters with sentences corresponding to less than 100 images per cluster were then dropped to form the final set of 24 labels as shown in Figure ?.

 %These images were annotated by 35 radiologists working on a cloud system using a structured report template. The resulting labels derived from these reports indicated many mentioned of tubes and lines and their placement issues in addition to anatomical findings.
 \vspace{-0.2in}
 \section{Classification of chest X-ray findings}
From the names of the labels available from AP Chest X-ray reports, we  can observe that the labels such as "left picc line with tip at the superior vena cava" and "left picc with tip at the cavoatrial junction," depict very similar appearance of these lines as shown in Figure~\ref{semanticoverlap} with the main difference being the position of the picc line (peripherally inserted central catheter) endpoint. In addition,  tubes and lines have a small footprint in the overall image due to their thin tubular structures. To ensure we are able to adequately distinguish between these finer granularity labels, we explored two different architectures for building the classifiers. The first architecture was an end-to-end deep learning network based on the DenseNet\cite{huang2017densely} which has proven to be very successful in classification problems for both scene image and chest x-ray imaging. %It alleviates the vanishing-gradient problem, strengthens feature propagation, encourages feature reuse, and substantially reduces the number of parameters.
In particular, a 121-layer DenseNet with weights initialized from a prior training on ImageNet\cite{imagenet}  was re-trained on the raw training images of our dataset and using 73 labels as output labels for the fully connected layer.  Our input images were resized to the ImageNet standard (224*224*3), and then centered using the ``caffe'' style of Keras's preprocess input function. The feature-maps of all  layers were combined and saved as a feature representation model in addition to supplying them as input to the fully connected layer for multi-way classification.
\begin{table}
\begin{tiny}
\centering
\begin{tabular}{|l|l|l|l|l|l|l|l}
\hline
Label	& Samples	&DFRF	&DenseNet\\
\hline
Averarge ROC	& &0.7&	0.69\\
Bibasal patchy opacities. &	223 &	0.68	&0.58\\
Bibasilar atelectasis, infection or aspiration.&	60&	0.71	&0.62\\
Bibasilar atelectasis.&	80&	0.63&	0.69\\
Bilateral pleural effusion.&	106	&0.78	&0.7\\
Blunting of bilateral costophrenic angles.&	118&	0.62	&0.7\\
Blunting of the left costophrenic angle.	&195	&0.64&	0.76\\
Blunting of the right costophrenic angle.	&108	&0.67&	0.7\\
Cardiac silhouette is enlarged.&	1271&	0.74&	0.8\\
Cardiac silhouette is mildly enlarged.	&246&	0.59&	0.58\\
Cephalization of the pulmonary vasculature.	&109	&0.71	&0.79\\
Diffuse bilateral opacities.	&264	&0.89&	0.9\\
Elevated left hemidiapragm.	&113	&0.53&	0.58\\
Elevated right hemidiaphram.&	163	&0.56&	0.75\\
Endotracheal tube present.&	130	&0.77	&0.76\\
Enlarged cardiac silhouette and diffuse parenchymal opacities&&&\\
which may represent volume overload/pulmonary edema.&	141	&0.67&	0.67\\
Enteric tube present.	&176	&0.79	&0.77\\
Enteric tube tip below the diaphragm. 	&185&	0.78&	0.77\\
Enteric tube with tip termination beyond the margin of the radiograph.&	103	&0.83	&0.72\\
ET tube in proper position.	&329	&0.84	&0.86\\
ET tube in trachea.	&186	&0.86&	0.81\\
Interstitial opacities bilaterally.&	93&	0.58&	0.75\\
Large body habitus.&	159&	0.87	&0.88\\
Left basal opacity.	&156	&0.66&	0.77\\
Left internal jugular line present.&	61&	0.61&	0.59\\
Left internal jugular line with tip at the cavoatrial junction.&	67	&0.8	&0.68\\
Left internal jugular line with tip overlying the superior vena cava.&	73&	0.56	&0.56\\
Left picc line present.	&247	&0.75	&0.8\\
Left picc line with tip overlying the superior vena cava.	&374	&0.62	&0.62\\
Left picc with tip at the cavoatrial junction.	&424	&0.71	&0.71\\
Left pleural effusion.	&77&	0.67	&0.66\\
\hline
\end{tabular}
\end{tiny}
\caption{Illustration of the label classes derived from the labeling process and the performance of DenseNet in AUC measure for the respective classes. Only 32 of the 73 derived labels are shown for brevity.}
\label{results}
\end{table}
In the second architecture, we formed a hybrid approach keeping the feature generation layers of deep learners and combining with an ensemble classifier. This was based on the rationale that advanced feature selection and explain capabilities of traditional ensemble classifiers may be more suitable for such higher grained label recognition. Specifically, we retained the feature representation model of the trained DenseNet and replaced the last fully-convolutional multi-class classification layer in DenseNet with an ensemble classifier.  We experimented with three separate boosting methods for random forests to address our inherent dataset imbalance, namely,  RUS \cite{seiffert2010rusboost}, Logit \cite{logitboost}, and Subspace \cite{yan2007model} boosting. We used deep trees, with a maximum number of tree splits equal to the size of our training set. We experimentally optimized our number of learning cycles to 1,000, and our learning rate to 0.1.
\vspace{-0.2in}
\section{Results}
\begin{figure}
	\centering
	\includegraphics[width=8cm]{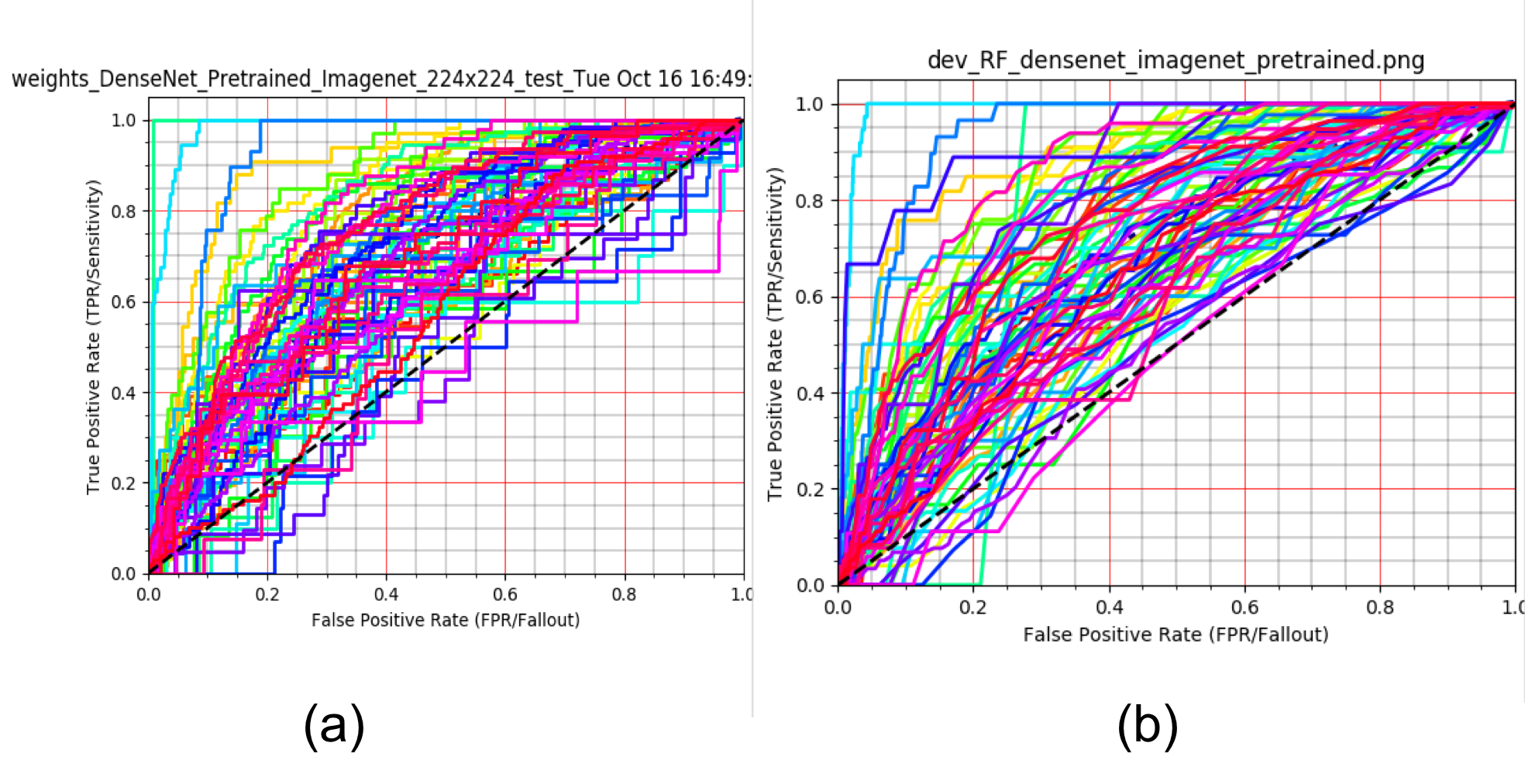}
\caption{Illustration of ROC curves for the 73 label dataset using (a) DenseNet (b) Deep ensemble classifier. }
	\label{roc}
\end{figure}

The experiments were performed with the newly labeled NIH dataset of 73 findings. %Figure~\ref{seventythree} shows the 30 most frequently occurring labels in the NIH re-annotated NIH dataset.
A total of 7942 images were retained corresponding to the 73 labels that had support of at least 50 images in the collection. A total of 6209 images were used for training, and 1733 were retained for validation and testing. First, we generated a baseline result using DenseNet directly on the dataset. The predicted labels were then used to plot the ROC curves and area under curve (AUC) was noted.  The resulting ROC curves and the average AUC are shown in Figure~\ref{roc}a.  In the next experiment, we used the hybrid learning model of DenseNet feature generator with the random forest classifier. The resulting ROC curves are shown in Figure~\ref{roc}b using 5-fold cross-validation on a 80-20 split of training and test data. From this figure, we see that the performance of the two networks are similar although with more training epochs and data, it is likely that DenseNet would eventually outperform the hybrid classifier (both achieved an average AUC of 0.7). The list findings and the AUC for the second classifier are shown in Table~\ref{results} (only the first 32 are shown of the 73 derived labels). We can also observe from the results in Table~\ref{results} that the models in general perform better for higher level abstraction labels if the number of images for training are also larger. We conclude from these results that it is possible to train classifiers to recognize finer distinction labels of AP chest X-rays. However,  the accuracy achieved still remains a function of the size of the labeled training datasets.
\vspace{-0.2in}
 \section{Conclusion}
% Below is an example of how to insert images. Delete the ``\vspace'' line,
% uncomment the preceding line ``\centerline...'' and replace ``imageX.ps''
% with a suitable PostScript file name.
% -------------------------------------------------------------------------
In this paper, we present a new chest X-ray benchmark database of 73 sentence-level findings seen in AP chest X-rays. We describe our method of obtaining these findings through a semi-automated ground truth generation process from crowdsourcing of clinical annotations. % To start a new column (but not a new page) and help balance the last-page
\bibliographystyle{IEEEbib}
%\bibliography{strings,refs}
\bibliography{isbibib}
\end{document}